\documentclass[conference,a4paper]{IEEEtran}
\IEEEoverridecommandlockouts
\usepackage{cite}
\usepackage{amsmath,amssymb,amsfonts}
\usepackage{algorithmic}
\usepackage{graphicx}
\usepackage{textcomp}
\usepackage{xcolor}
\usepackage{verbatim}
\usepackage{booktabs}
\usepackage[table]{xcolor}
\def\BibTeX{{\rm B\kern-.05em{\sc i\kern-.025em b}\kern-.08em
    T\kern-.1667em\lower.7ex\hbox{E}\kern-.125emX}}
\begin{document}
\hyphenation{Standardisation 
management
O-RAN
Learning-to-Optimise}

%*********************AEIT COPYRIGHT****************************
\makeatletter
\def\ps@IEEEtitlepagestyle{%
  \def\@oddfoot{\mycopyrightnotice}%
  \def\@evenfoot{}%
}
\def\mycopyrightnotice{%
  {\footnotesize 978-88-87237-64-1 \copyright 2026 AEIT \hfill} 
}
\makeatother
%***************************************************************

\title{Learning-to-Optimize as the Missing Architectural Layer of AI-Native Networks\\
\thanks{This work has been carried out in the framework of a project between the Ministry of "Imprese e Made in Italy" and Fondazione Ugo Bordoni.}
}

\author{\IEEEauthorblockN{Giambattista Amati,Federica Mangiatordi, Pierpaolo Salvo, Emiliano Pallotti, Simone Angelini}
\IEEEauthorblockA{\textit{Fondazione Ugo Bordoni } \\
Viale del Policlinico 147,Rome, Italy \\
\{gamati, fmangiatordi, psalvo, epallotti, sangelini\}@fub.it}
}

\maketitle

\begin{abstract}

Artificial Intelligence (AI) is becoming a fundamental design principle of future AI-native communication networks, enabling autonomous resource management, adaptive control, and zero-touch network operation. While current AI-native architectures increasingly embed intelligence across network functions, they provide little guidance on how optimisation knowledge should be systematically generated, transferred, and exploited by AI models.
This paper argues that the Learning-to-Optimize (L2O) represents the missing architectural layer between optimisation and AI-native intelligence. Rather than viewing optimisation merely as an online decision engine, the proposed paradigm redefines optimisation algorithms as offline knowledge generators that produce high-quality supervisory information for neural surrogate models. The resulting models inherit optimisation expertise while enabling low-latency runtime inference suitable for dynamic network environments.
A generic four-stage L2O workflow is introduced, comprising optimisation, knowledge generation, surrogate learning, and runtime inference. Unlike existing Learning-to-Optimize approaches, which primarily focus on algorithm acceleration, the proposed framework establishes L2O as an architectural abstraction applicable across heterogeneous communication and computing systems.
The proposed paradigm is illustrated by an NR-V2X relay-selection problem, in which optimisation-generated solutions from a Mixed-Integer Linear Programming (MILP) solver are used to train a Graph Neural Network that can reproduce near-optimal decisions in real time. The presented perspective positions Learning-to-Optimize as a key architectural enabler for future AI-native networks.

\end{abstract}


\section{Introduction}

Artificial Intelligence (AI) is increasingly recognised as a foundational technological paradigm for the design and operation of next-generation communication infrastructures. Whereas earlier generations of mobile networks primarily employed AI to improve isolated management and optimisation functions, AI-native 6G architectures conceptualise intelligence as an intrinsic, pervasive attribute of the overall system design. Rather than functioning as an exogenous optimisation tool, AI is envisaged to be deeply integrated across the communication, computation, control, and service-orchestration layers, thereby enabling autonomous, adaptive, and context-aware network behaviour, as well as holistic lifecycle management.

This vision is reflected in the ongoing evolution of international standardisation and research initiatives.  Standardisation bodies, including 3GPP, ETSI \cite{tregpp23288}, and the O-RAN Alliance \cite{ORAN1}, are progressively incorporating artificial intelligence (AI) into network management and control frameworks. In parallel, major European research initiatives—such as 6GARROW, AIORA, and 6G-GOALS—identify distributed AI agents, digital twins, semantic communications, and AI-driven orchestration as fundamental building blocks of next-generation communication systems \cite{Matera2024, 6garrow2025,6garrow_overview, AIORA, Goals2025, Toumi2025}. Together, these efforts are driving the transition from AI-assisted to genuinely AI-native network architectures.

The increasing intelligence of communication systems is accompanied by a growing optimisation complexity. Core network functions—including resource allocation, topology management, routing, relay selection, service placement, task offloading, spectrum coordination, power control, and cloud-edge orchestration—are naturally formulated as constrained optimisation problems\cite{11455199,TYCHOGIORGOS201432,Patil2024,10239348,wang2018}. Mathematical programming, graph optimisation, and metaheuristic algorithms have traditionally provided highly effective solutions, often achieving globally optimal or near-optimal performance. However, their computational complexity rapidly becomes prohibitive as network scale and problem dimensionality increase, making online optimisation incompatible with the stringent latency requirements of AI-native network control.

To overcome these limitations, the research community has increasingly investigated \emph{Learning-to-Optimise} (L2O), a family of techniques that learn to approximate optimisation procedures using machine learning. Rather than solving each optimisation problem online, L2O uses offline-accumulated optimisation experience to train surrogate models that produce optimisation-quality decisions with dramatically lower inference latency \cite{Tang2024,Bengio2021,10736694,11268640,Mazyavkina2021}.

To address this limitation, we introduce the Learning-to-Optimise (L2O) Layer, a new architectural abstraction bridging optimisation engines and AI-native controllers. 

To illustrate the proposed paradigm, we consider an NR-V2X relay-selection problem in which optimal communication topologies, generated by a Mixed-Integer Linear Programming (MILP) solver, are used to train a Graph Neural Network (GNN) that reproduces near-optimal relay-selection decisions in real time. Although relay selection serves as an illustrative example, the proposed Learning-to-Optimise layer naturally generalises to a broad class of communication and computing optimisation problems.

The main contributions of this work are as follows:

\begin{itemize}

\item We analyse the relationship between AI-native network architectures and Learning-to-Optimise, identifying the absence of an explicit architectural mechanism connecting optimisation and AI.

\item We introduce the Learning-to-Optimise Layer as a novel architectural abstraction that systematically transforms optimisation knowledge into deployable AI models.

\item We define a generic four-stage Learning-to-Optimise workflow comprising optimisation, knowledge generation, surrogate learning, and runtime inference.

\item We demonstrate the proposed paradigm through an optimisation-guided Graph Neural Network for relay selection in NR-V2X networks, illustrating how optimisation-generated knowledge enables near-optimal real-time network control.

\end{itemize}

%%%%%%%%%%%%%%%%%%%%%%%%%%%%%%%%%%%%%%%%%%%%%%%%%%%%%%%%%%%%%%%%%%%%%%%%%%%%%%%
%%%%%%%%%%%%%%%%%%%%%%%%%%%%%%%%%%%%%%%%%%%%%%%%%%%%%%%%%%%%%%%%%%%%%%%%%%%%%%%
\begin{figure*}[h]
    \centering
    \includegraphics[width=1.0\linewidth]{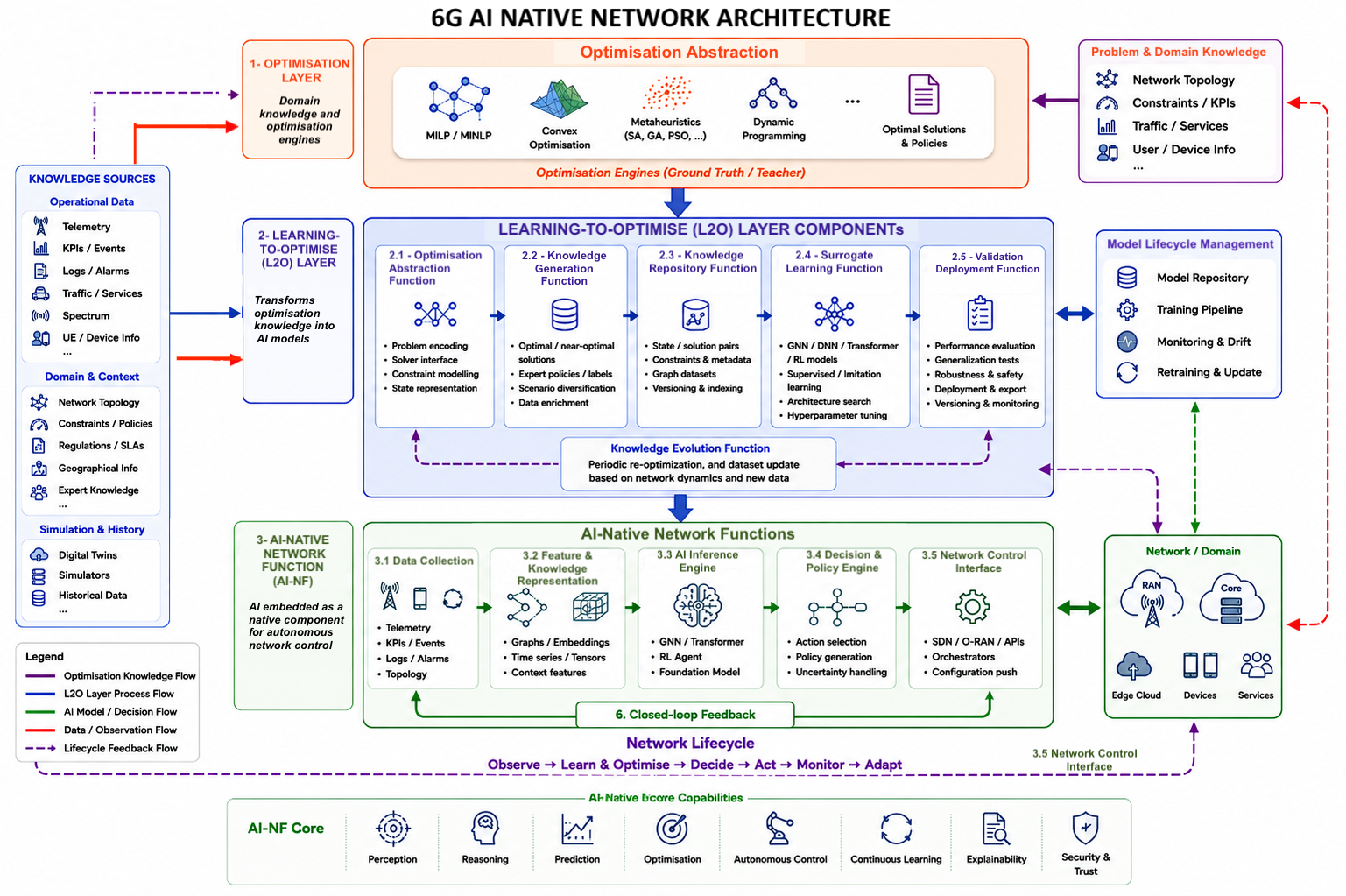}
    \caption{Proposed AI-native network architecture incorporating the Learning-to-Optimize (L2O) Layer as the missing architectural bridge between optimisation engines and AI-native network functions}
    \label{fig:l2o_architecture}
\end{figure*}

\section{Architectural Analysis and Research Gap}
This section compares existing AI-native architectures and
Learning-to-Optimise methodologies in order to identify the
architectural functionality that remains unsupported by current
frameworks.

\subsection{Existing AI-Native Functional Architectures}

\textcolor{black}{The \emph{native AI} paradigm was identified in early 6G vision papers, where native AI was listed among the architectural features of 6G together with lite network, soft network, and native security~\cite{Liu2020Vision}. This vision introduced the idea that intelligence should become a built-in property of future mobile networks rather than an external add-on.}

Recent AI-native architectures, including those proposed within 3GPP, O-RAN, 6GARROW, AIORA and 6G-GOALS, converge toward a common functional organisation in which AI is embedded throughout
network management and control \cite{Matera2024,6garrow2025,6garrow_overview,AIORA,Goals2025,Toumi2025}.

\textcolor{black}{\textcolor{black}The AI-native User-Centric Network architecture follows the same direction by introducing a cloud-based control plane, an edge/distributed user plane, and an intelligence plane for data collection, AI model training and inference, and AI algorithm generation and tuning~\cite{ChenLi2022UCN}.}

\textcolor{black}{UNITY-6G further extends this direction with a distributed AI-native management-and-orchestration layer, a digital twin, semantic communication, and an integrated architecture spanning Open RAN, edge-cloud, core, transport, and non-terrestrial domains~\cite{UNITY6G2025}. Likewise, the foundation model-based cloud-edge-end collaboration framework strengthens the native-AI vision through pre-trained foundation-model agents, advanced retrieval-augmented generation, expert knowledge integration, and hierarchical multi-agent collaboration~\cite{ChenXiang2025}.}

\textcolor{black}{Among these approaches, the model-dependency framework proposed by Toumi and Dimitrovski~\cite{Toumi2025} represents the closest attempt to structure the relationship between AI/ML models and their supporting data sources. Their solution introduces a repository that tracks dependencies among models, Digital Twins, and Transfer Learning processes, enabling degradation events in a base model to trigger corrective actions in dependent models. However, this framework governs the lifecycle and interdependencies of AI/ML models that are assumed to pre-exist, rather than defining how such models should be systematically generated from an optimisation process. The origin of the supervisory knowledge itself therefore remains outside the scope of the architecture, which is precisely the aspect addressed by the Learning-to-Optimise Layer proposed in this paper.}

Despite differences in implementation, these architectures consistently
introduce functional entities for network observability, distributed AI
inference, AI lifecycle management, Digital Twins, and intelligent
orchestration. Their primary contribution is to define how AI models
are deployed, executed, monitored, and updated during network
operation.

However, they generally consider AI models as architectural assets
that are already available. The generation of optimisation-derived AI
models is typically assumed to occur outside the architectural
framework and is therefore not explicitly addressed.

\subsection{Learning-to-Optimise: An Algorithmic Perspective}

Learning-to-Optimise has emerged as an effective paradigm for
approximating computationally expensive optimisation procedures
through machine learning
\cite{Tang2024,Bengio2021,Mazyavkina2021}.

Instead of solving optimisation problems online, L2O relies on
offline-generated optimisation solutions to train surrogate models
capable of producing near-optimal decisions with significantly lower
runtime complexity.

Existing L2O approaches, however, primarily focus on improving the
efficiency of optimisation algorithms or replacing them with learned
approximations \cite{chen2021}. Consequently, they address optimisation from an
algorithmic perspective rather than from an architectural one.

\subsection{Research Gap}

The previous analysis highlights a clear separation between the two
research directions.

AI-native architectures define how AI models are managed throughout
their operational lifecycle but do not specify how optimisation
knowledge should be transformed into deployable AI models.

Conversely, Learning-to-Optimise explains how optimisation expertise
can be transferred to machine learning models but does not define how
this process should be integrated into an AI-native architecture.

\textcolor{black}{Even approaches that explicitly model relationships between AI artefacts, such as~\cite{Toumi2025}, address dependency management among pre-existing models rather than the systematic generation of optimisation-derived knowledge, confirming that the two research directions remain structurally disconnected.}

Existing AI-native architectures manage the lifecycle of AI
models, whereas conventional Learning-to-Optimise methods focus on generating surrogate models from optimisation.
Neither defines an architectural lifecycle for optimisation-derived knowledge.

Consequently, no architectural component currently exists to systematically generate, manage, and reuse optimisation knowledge across the AI lifecycle. This observation raises the following architectural question:

\begin{quote}
\emph{How can optimisation expertise be systematically
generated, managed, and transformed into reusable AI models within AI-native network architectures?}
\end{quote}

To address this limitation, we introduce the proposed
Learning-to-Optimise Layer, which establishes a functional bridge
between optimisation engines and AI-native control functions.

\begin{table*}[h]
\centering
\caption{Comparison between conventional AI-native architectures, conventional Learning-to-Optimise (L2O), and the proposed Learning-to-Optimise Layer.}
\label{tab:l2o_gap}
\renewcommand{\arraystretch}{1.2}
\begin{tabular}{p{3.1cm}p{3.5cm}p{3.4cm}p{4.5cm}}
\hline
\textbf{Architectural capability} &
\textbf{AI-native architectures} &
\textbf{Conventional L2O} &
\textbf{Proposed L2O Layer} \\
\hline

Primary objective &
AI lifecycle management &
Optimisation acceleration &
Optimisation knowledge lifecycle \\

AI model generation &
External / assumed &
Generated from optimisation &
Generated from optimisation \\

Role of optimisation &
Runtime optimisation support &
Optimisation oracle &
Persistent knowledge generation \\

Knowledge management &
Model repository &
Optimisation datasets &
Optimisation knowledge repository \\

Lifecycle focus &
Deployment and retraining &
Offline surrogate learning &
Knowledge generation, learning, deployment, and evolution \\

Supervision source &
Training datasets &
Optimisation labels &
Optimisation oracle and knowledge repository \\

Architectural role &
AI management function &
Algorithmic workflow &
Native architectural component \\

Runtime decision making &
AI inference &
Standalone surrogate inference &
AI-native optimisation-aware inference \\

\hline
\end{tabular}
\end{table*}

%%%%%%%%%%%%%%%%%%%%%%%%%%%%%%%%%%%%%%%%%%%%%%%%%%%%%%%%%%%%%%%%%%%%%%%%%%%%%%%

\section{The Learning-to-Optimise Layer: A Functional Architectural Component}

Figure~\ref{fig:l2o_architecture} illustrates the proposed Learning-to-Optimise Layer and its positioning within an AI-native architecture. The layer serves as a persistent architectural knowledge plane that bridges heterogeneous optimisation engines and AI-native control functions by managing optimisation-derived knowledge.

\subsection{Architectural Role}
The Learning-to-Optimise Layer operates as an intermediate
knowledge plane between the network infrastructure and
AI-native control functions. 
\textcolor{black}{It can be deployed as a logically
centralised but physically distributed service, with its
functional components instantiated across  network, edge, or
cloud resources according to latency, scalability, and
data-locality requirements.}
Rather than computing runtime control decisions directly, the
layer transforms heterogeneous network observations into
persistent optimisation-derived knowledge that supports
AI-native decision making throughout the network lifecycle.
The layer acquires observations from multiple sources,
including:

\begin{itemize}

\item \textit{Physical layer:} channel quality measurements,
interference statistics, beam management information,
spectrum occupancy, and radio resource measurements;

\item \textit{User plane:} traffic demand, latency, throughput,
packet loss, QoS indicators, and service flows;

\item \textit{Control plane:} routing configurations, topology
evolution, slicing policies, and resource allocation;

\item \textit{Knowledge sources:} Digital Twins, simulators,
and historical operational repositories describing representative network conditions.
\end{itemize} These heterogeneous observations are abstracted into
optimisation problems whose solutions constitute reusable optimisation knowledge. Such knowledge is persistently stored,
continuously enriched, and shared across multiple AI-native functions through the Learning-to-Optimise Layer.

\subsection{Functional Components}

\subsubsection{Optimization Abstraction Function}
The optimisation engines are not part of the Learning-to-Optimise Layer itself; instead, they are external computational services accessed through the Optimisation Abstraction Function.
The Optimisation Abstraction Function provides a technology-independent interface to heterogeneous optimisation engines, including MILP, convex optimisation, graph algorithms, and metaheuristics. It abstracts optimisation-specific details and exposes a unified representation of optimisation problems and solutions to the remaining functional components of the Learning-to-Optimize Layer.

\subsubsection{Knowledge Generation Function}

Representative network states are processed by the optimisation engines to generate optimal or near-optimal decisions. These optimisation outcomes constitute reusable expert knowledge describing optimal network behaviour under representative operating conditions.
\textcolor{black}{To enable effective surrogate learning, the generated knowledge should adequately span the variability of representative network states, ensuring sufficient coverage of the target operating conditions.}

\subsubsection{Knowledge Repository Function}
Optimisation-derived knowledge is persistently stored alongside the corresponding network state, optimisation metadata, constraints, objective function, solver information, and decision labels.
The repository serves as a reusable knowledge base for surrogate learning, benchmarking, continual learning, and future optimisation campaigns.

\subsubsection{Surrogate Learning Function}

The optimisation knowledge stored in the repository is used to train surrogate AI models that learn the mapping between network states and optimisation decisions. Depending on the target application, surrogate models may be implemented using Graph Neural Networks, Deep Neural Networks, Transformers, Reinforcement Learning agents, or other AI architectures.
Rather than reproducing the optimisation algorithm itself, the surrogate learns the optimisation-induced decision policy, preserving the mapping between network states and decisions while exploiting its own internal representation.

\subsubsection{Model Validation and Deployment Function}
Before deployment, surrogate models are validated against the reference optimisation solutions to verify decision accuracy, constraint satisfaction, and robustness under previously unseen operating conditions.
\textcolor{black}{Once validated, the models are deployed as AI-native network functions, providing optimisation-quality decisions through low-latency runtime inference.}

\subsubsection{Knowledge Evolution Function}

Network conditions continuously evolve due to traffic variations, topology modifications, and environmental changes. The Knowledge Evolution Function periodically triggers new optimisation campaigns to enrich the knowledge repository and retrain surrogate models whenever significant changes in the operating environment are detected.
\textcolor{black}{Triggering policies may rely on performance degradation, topology or traffic changes, or the availability of newly collected representative network states that extend the coverage of the target operating regime.}

This continuous update mechanism enables the Learning-to-Optimise Layer to maintain the relevance of optimisation knowledge throughout the network's operational lifecycle.

\subsection{Architectural Workflow}
The interaction among the functional components is illustrated
in Fig.~\ref{fig:l2o_architecture}.

Network observations collected from the infrastructure and
Digital Twins are first abstracted into optimisation problems
and processed by heterogeneous optimisation engines.
The resulting optimisation knowledge is stored within the
Knowledge Repository, exploited to train surrogate AI models,
validated before deployment, and continuously enriched through
periodic optimisation campaigns as network conditions evolve.

Unlike conventional Learning-to-Optimise pipelines, this
workflow describes the interaction among persistent
architectural functions rather than the execution of a single
optimisation algorithm. \textcolor{black}{Consequently, the proposed
Learning-to-Optimise Layer provides a reusable architectural
abstraction that is independent of the underlying optimisation
problem and applicable to heterogeneous communication and
computing systems.
}

%%%%%%%%%%%%%%%
%-----
\section{Architectural Instantiation: NR-V2X Relay Selection}
\begin{figure}[h!]
    \centering
    \includegraphics[width=0.9\linewidth]{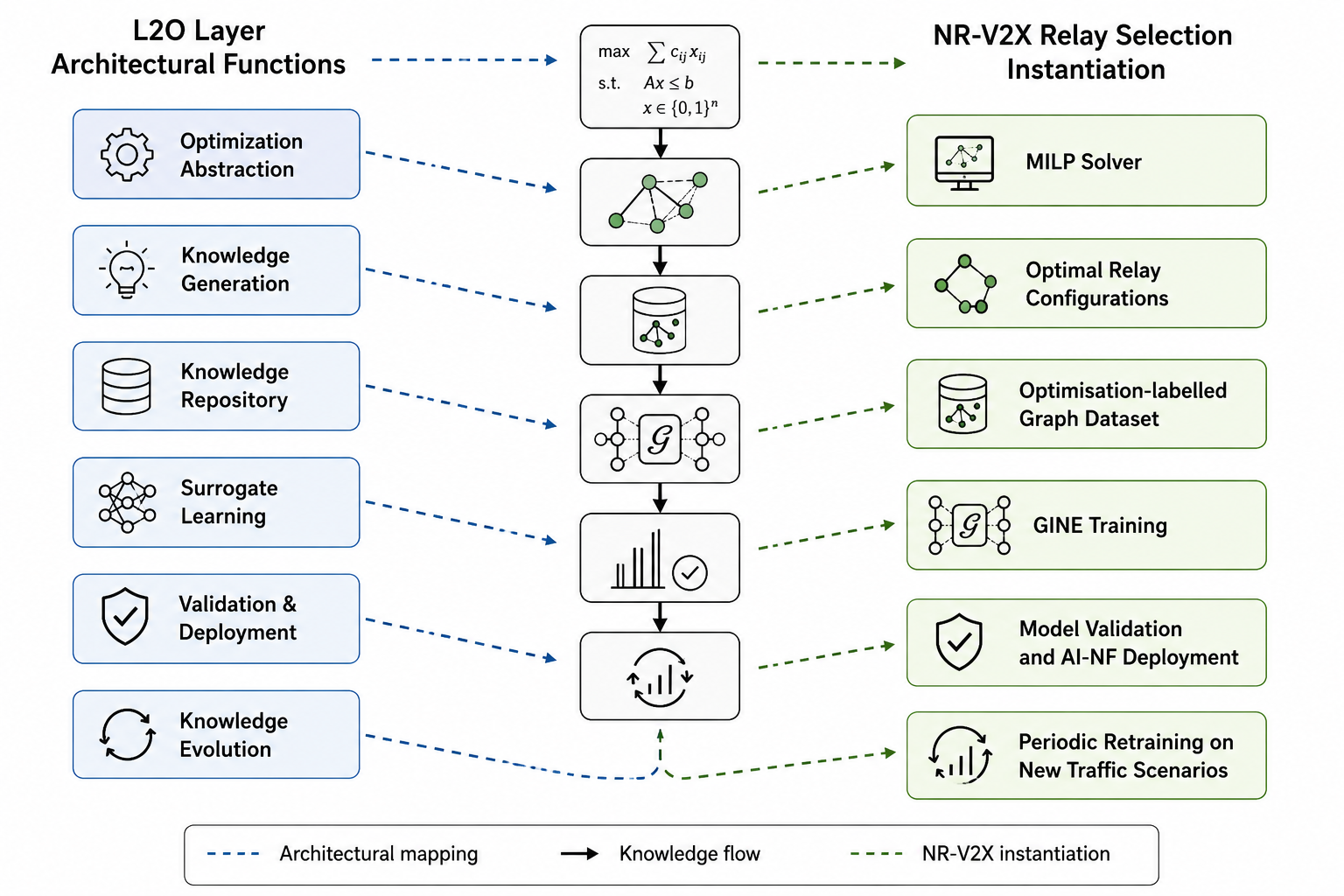}
    \caption{Functional architecture of the proposed Learning-to-Optimize (L2O) Layer and its instantiation in the NR-V2X relay-selection case study}
    \label{fig:placeholder}
\end{figure}

To illustrate a concrete instantiation of the proposed Learning-to-Optimise (L2O) Layer, we consider the relay-selection problem in dense NR-V2X networks. Rather than evaluating a specific optimisation algorithm, this case study demonstrates how the functional components introduced in Section~III interact to systematically transform optimisation expertise into an AI-native network function. The overall architectural instantiation is illustrated in Fig.~\ref{fig:placeholder}, while Table~\ref{tab:instantiation} summarises the correspondence between the proposed L2O functions and their implementation in the considered use case.

Reliable uplink connectivity is a fundamental requirement for Connected and Automated Vehicles (CAVs), particularly in dense urban environments where sparse Road-Side Units (RSUs), non-line-of-sight propagation, and rapidly changing topologies often preclude direct access to infrastructure. Relay selection can therefore be formulated as a constrained combinatorial optimisation problem whose objective is to maximise end-to-end connectivity while satisfying routing, flow conservation, and capacity constraints.

\subsection{Instantiation of the Functional Components}

The \emph{Optimization Abstraction Function} is instantiated through a Mixed-Integer Linear Programming (MILP) solver previously developed by the authors. The optimisation engine computes globally optimal relay configurations for representative NR-V2X communication graphs. Although the MILP provides optimal decisions, its computational complexity makes online execution unsuitable for real-time vehicular scenarios.

Representative network states are generated through an integrated OSM--SUMO--GEMV2 simulation framework. Each snapshot is represented as a directed communication graph whose nodes correspond to CAVs and RSUs, while edges describe feasible wireless links enriched with physical-layer information, including signal-to-noise ratio and achievable Shannon capacity.

For every representative network state, the optimisation solver produces the corresponding optimal relay configuration, thereby instantiating the \emph{Knowledge Generation Function}. These optimisation-derived decisions constitute the optimisation knowledge generated by the proposed L2O Layer.

The resulting optimisation-labelled communication graphs populate the \emph{Knowledge Repository Function}. Each repository entry consists of a graph representation paired with the corresponding optimisation-derived relay decisions, forming the optimisation-derived dataset used for surrogate learning and future knowledge evolution.

The \emph{Surrogate Learning Function} is instantiated through an edge-aware Graph Isomorphism Network (GINE), which learns the mapping between graph-structured network states and the optimisation-derived relay decisions generated by the MILP oracle. Rather than reproducing the optimisation algorithm itself, the surrogate learns the optimisation-induced decision policy.

The \emph{Validation and Deployment Function} evaluates the surrogate model against the optimisation oracle with respect to prediction accuracy, constraint satisfaction, and generalisation capability. Once validated, the model is integrated into the AI-native relay-selection function, enabling real-time relay selection through low-latency inference while removing the optimisation solver from the runtime control loop.

Finally, the \emph{Knowledge Evolution Function} is realised by periodically generating additional representative traffic scenarios, solving the corresponding optimisation problems, and enriching the repository with new optimisation knowledge. This enables the surrogate model to be retrained as network conditions evolve without modifying the remaining architectural components.
\subsection{Experimental Results}

Experimental results demonstrate that the GINE reproduces the MILP relay decisions with approximately 96\% link-level accuracy while maintaining inference latency below 5~ms for all evaluated graph instances \cite{Amati2026ICCSPA,11604994}. Furthermore, when integrated into a hybrid optimisation framework, the learned surrogate effectively prunes the MILP search space, enabling the recovery of MILP-equivalent solutions with substantially reduced execution time.

The correspondence between the functional components of the proposed Learning-to-Optimise Layer and their NR-V2X implementation is summarised in Table~\ref{tab:instantiation}. Although instantiated here through relay selection, the same architectural workflow is directly applicable to a broad class of AI-native network functions, including routing, resource allocation, spectrum coordination, topology optimisation, service placement, task offloading, and cloud-edge orchestration.

Consequently, the proposed Learning-to-Optimise Layer remains independent of both the optimisation methodology and the surrogate AI model employed, providing a reusable architectural mechanism for systematically transforming optimisation knowledge into deployable AI-native network intelligence.

\begin{table}[t]
\centering
\caption{Instantiation of the proposed Learning-to-Optimize Layer in the NR-V2X relay-selection use case.}
\label{tab:instantiation}
\renewcommand{\arraystretch}{1.15}
\begin{tabular}{ll}
\hline
\textbf{L2O Function} & \textbf{NR-V2X Instantiation} \\
\hline
Optimization Abstraction & MILP solver \\
Knowledge Generation & Optimal relay configurations \\
Knowledge Repository & Optimisation-labelled graph dataset \\
Surrogate Learning & GINE training \\
Validation \& Deployment & Model validation and AI-NF deployment \\
Knowledge Evolution & Periodic retraining on new traffic scenarios \\
\hline
\end{tabular}
\end{table}

\section{Conclusions}

This paper introduced the \emph{Learning-to-Optimise (L2O) Layer} as a functional architectural component for AI-native networks. Unlike conventional Learning-to-Optimise approaches, which primarily focus on accelerating optimisation algorithms, the proposed architecture explicitly integrates optimisation-derived knowledge into the AI-native lifecycle through a reusable architectural framework.

The proposed L2O Layer defines a general workflow comprising optimisation abstraction, knowledge generation, knowledge management, surrogate learning, model deployment, and continuous knowledge evolution. This functional decomposition enables the systematic generation of optimisation knowledge offline and its subsequent exploitation by AI-native control functions via low-latency surrogate inference.
The NR-V2X relay-selection case study illustrated how the proposed architecture can be instantiated in practice by combining an MILP optimisation engine with a Graph Neural Network surrogate. Although demonstrated for relay selection, the same architectural principles naturally extend to a wide range of communication and computing problems, including routing, spectrum coordination, resource allocation, topology optimisation, service placement, and cloud-edge orchestration.

Future work will investigate constraint-aware surrogate learning, continual knowledge evolution strategies, and Digital Twin-assisted optimisation, as well as the integration of the proposed Learning-to-Optimise Layer within emerging AI-native 6G network architectures.
\section*{Acknowledgement} %\textsuperscript{*}Note: Sub-titles are not captured for https://ieeexplore.ieee.org  and
The work has been carried out in the framework of the Spectrum Sharing project between the Ministry of Enterprises and Made in Italy (MIMIT) and Fondazione Ugo Bordoni.

\bibliographystyle{IEEEtran}
\bibliography{ References.bib}

\end{document}